# USE SQUARE ROOT AFFINITY TO REGRESS LABEL IN SEMANTIC SEGMENTATION

*Lumeng Cao*

clm@mail.ustc.edu.cn

## ABSTRACT

Semantic segmentation is a basic but non-trivial task in computer vision. Many previous work focus on utilizing affinity patterns to enhance segmentation networks. Most of these studies use the affinity matrix as a kind of feature fusion weights, which is part of modules embedded in the network, such as attention models and non-local models. In this paper, we associate affinity matrix with labels, exploiting the affinity in a supervised way. Specifically, we utilize the label to generate a multi-scale label affinity matrix as a structural supervision, and we use a square root kernel to compute a non-local affinity matrix on output layers With such two affinities, we define a novel loss called Affinity Regression loss (AR loss), which can be an auxiliary loss providing pair-wise similarity penalty. Our model is easy to train and adds little computational burden without run-time inference. Extensive experiments on NYUv2 dataset and Cityscapes dataset demonstrate that our proposed method is sufficient in promoting semantic segmentation networks.

*Index Terms*— Convolutional neural network, Semantic segmentation, Affinity

## 1. INTRODUCTION

Semantic segmentation is a challenging task in computer vision, which is applied in various fields such as autonomous driving, robots, satellites, agriculture, medical diagnosing, etc. It is a dense classification task which aims to provide per-pixel classification of an image. Thanks to the rapid development of convolutional neural network technology, many semantic segmentation neural networks are developed. For example, FCN [1] used a convolutional layer to replace the fully connected layer, making the neural network adaptable to any input size. Deeplab [2], PSPNet [3] adapted spatial pyramid pooling block to extract features at different scales, and then merged the features to capture contextual information of different sizes. For a long time, researchers focused on feature-reuse method and attention mechanisms to design segmentation networks. [4, 5, 6] used residual and dense skip connections to aggregate and reuse features in various levels to make semantic segmentation more accurate and make the gradient easier to pass short or long forward. Attention models [7, 8, 9] and non-local models [10, 11] make up for the local limitation of convolutional kernels, and can capture long range dependency. Recent studies show the importance of pixel grouping [12, 13, 14, 15]. Zhong et al [12] proposed that semantic segmentation can be disentangled into two subtasks: explicit pixel-wise prediction and implicit pixel grouping. Yu et al [13] used labels to model prior knowledge within and between classes to guide the learning of context information. Ke et al [14] proposed an adaptive affinity fields (AAF) to capture and match the semantic relations between neighbouring pixels in the label space. Jiang et al [15] proposed a diffusion branch composed of a seed subbranch for score map and a similarity subbranch for pixel-wise similarities. CRFs [16, 17, 18, 2] methods are used in semantic segmentation to leverage context information to optimize network output, which is a statistical technique to group similarity pixels and refine score maps by energy functions. Many previous CRFs are post-processing of network output. Vemulapalli et al [17] and Chandra et al [18] introduced Gaussian Conditional Random Fields in CNNs and achieve good results.

The advanced technologies in semantic segmentations such as attention mechanisms, non-local models, affinity propagation mechanisms require high computational overhead [7, 11, 19]. It is mainly because that the size of affinity matrix $X^TX$ ($X \in R^{C \times HW}$) is too large. in many practical applications, we need a light fast network to deal with requirements like GPU limitations and real-time demands. In addition to designing light and efficient networks [20, 21, 22], it is also very meaningful to explore ways to improve original semantic segmentation networks without adding extra computational burden. We notice that in the methodology of pixel grouping, it is very important to model pair-wise relationships. Inspired by the previous work [16, 15, 11] in utilizing binary computation of positions and feature vectors, we adapt the label affinity [14, 13] and connected it with score maps(feature maps before SoftMax) as a penalty for similarity. Unlike previous works [14, 13], We model a non-local affinity guidance in a regression way. Besides, we design a square root kernel for affinity computation, making regression penalty mathematically interpretable. Furthermore, we adapt the spatial pyramid pooling module to aggregate different scale information and reduce computational burden. Experiments on NYUv2 and Cityscapes datasets demonstrate the effectiveness of our model. Our main contributions can be summarized into three aspects:

- We propose a novel auxiliary loss function providing pair-wise supervision that cross-entropy loss cannot provide in semantic segmentation tasks.
- We model the pair-wise loss in a regression way to help solve classification-based problems and it is mathematically interpretable.
- Our model increases little computational burden or GPU memory without run-time inference.

## 2. RELATED WORK

Cross-Entropy CE loss is a widely used unary loss function in classification tasks. In semantic segmentation, basic CE loss is mainly used and there are some other supplements. OHEM algorithm increases the weight of misclassified samples. Focal loss [23] reduces the weight of easy-to-classify samples to make the model focus more on difficult-to-classify samples. IOU loss and Teversky loss provide coarse scale supervisions. The cross-entropy loss in semantic segmentation is:

$$H(p) = -\sum_c y_c \log p_c \quad (1)$$

where $y_c$ is 0 except for 1 position on a single pixel, which is a unary supervision.

Pair-wise Modeling Conditional random fields additionally consider the binary relationship between prediction points by energy function. A typical CRF model [2] can be written as:

$$E(X) = \sum_i \theta_i(x_i) + \sum_{ij} \theta_{ij}(x_i, x_j) \quad (2)$$

where $O_i$ is a unary potential function with probability outputs from DCNNs and $O_{ij}$ is pair-wise potential function computing the similarity of color intensity and positions between two points, aiming to force pixels with similar color and position to have similar labels.

Non-local models [10, 11] and many attention models [9, 8, 7] also focus on utilizing the binary computation of ($x_i$, $x_j$), where ($x_i$, $x_j$) are feature map vectors. In these attempts, SoftMax transformation is used on the rows of affinity matrix to generate feature fusion weights in order to capture long range relationship in spatial wise and channel wise. Naturally, affinity matrix occupies a core position in pair-wise modeling. To model the pair-wise relationships, formally, suppose $x_i$, $x_j$ are the feature vectors of the $ith$ and $jth$ positions, we can define their similarity $s(x_i, x_j)$ through some functions such as inner product $x_i^T x_j$, L1 distance $||x_i - x_j||$, exponential function $e^{-s(x_i, x_j)}$ and other mathematical forms. In our work, we compute square root operation before dot product functions: $\sqrt{x_i^T}\sqrt{x_j}$ and $x_i$, $x_j$ are SoftMax normalized. This form can preserve probability information and make the dot-product affinity pattern being Lipschitz functions[24].

Spatial Pyramid Pooling He et al [25] successfully adapted spatial pyramid pooling into object detection task, Chen et al [2] used dilated convolutions of different sampling weights as a substitute of pooling layers and designed an Atrous Spatial Pyramid Pooling layer (ASPP) to account for multiple scales. PSPNet [3] conducted spatial pyramid pooling after a specific layer to embed context features of different scales into the networks. Meng et all [11] modified spatial pyramid pooling module to fit non-local models in order to reduce computational burden. We follow the multi-scale pooling policy in building non-local affinity patterns.

## 3. METHOD

### 3.1. Label Affinity Matrix

The label of semantic segmentation is a single channel picture with discrete values, and each value represents one class. The operation for generating label affinity matrix is similar to self-attention operations in many vision tasks: given a label picture of size 1 x W x H, flatten the label to one dimension to get a 1 x WH vector L, then we get an affinity matrix M of size *WH x WH*, which satisfies:

$$M_{i,j} = \begin{cases} 0 & \text{if } L[i] \neq L[j] \text{ for i,j} \\ 1 & \text{if } L[i] = L[j] \text{ for i,j} \end{cases} \quad (3)$$

The M matrix is very large, for example, if the size ofa label picture is 100 x 100, then L is of 1 x 10000 length andM is of 10000 x 10000 size. Besides, M is sparse in scenes with many distinctive labels. Itis hard to model a large sparse matrix. We adapt the PSPModule, using a symmetry psp module to make the computation workable. We use nearest neighbor interpolation to down sample the label into sizes of 12 x 12, 6 x 6, and then flatten the label and concatenate them together to get a 1 x 180 sized L. Figure.1 demonstrates the process.

### 3.2. Square Root Affinity

A typical non-local affinity embedded in attention models or non-local models is like this:
Given an input feature $X \in \mathcal{R}^{C \times H \times W}$, where C,H,W denote the the dimension of channel, spatial height and width respectively.
Three different 1 x 1 convolutions $W_\phi$, $W_s$ and $W_Y$ map X to different embeddings, Y G Rc $\phi \in \mathcal{R}^{\hat{C} \times H \times W}, \theta \in \mathcal{R}^{\hat{C} \times H \times W}$, $\gamma \in \mathcal{R}^{\hat{C} \times H \times W}$ as: $\phi = W_\phi(X), \theta = W_\theta(X), \gamma = W_\gamma(X)$, $\hat{C}$ represents the channel of new embeddings. Then three embeddings are flattened to size C x L, where $L = H \times W$. The affinity matrix $A \in \mathcal{R}^{L \times L}$is thus calculated as:

$$A = \phi^T \times \theta \quad (4)$$

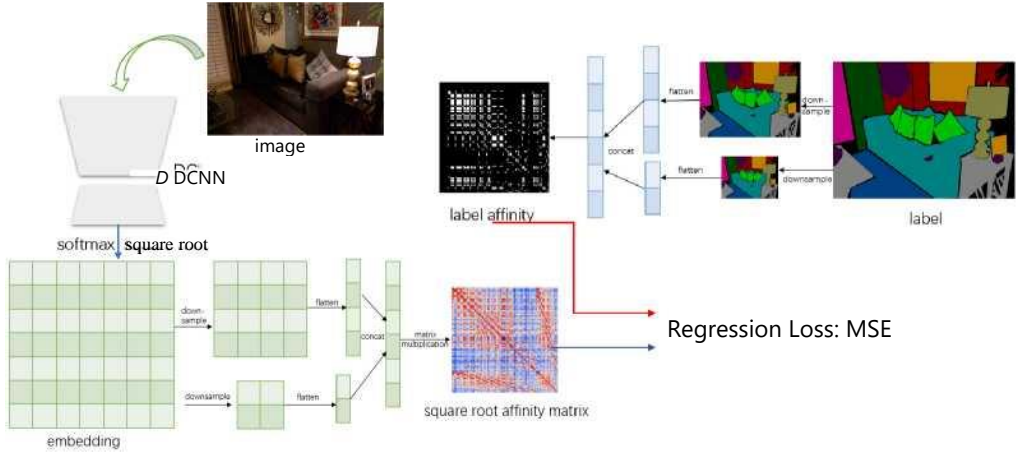

Figure.1. Architecture of the proposed affinity regression loss model

The next step is some normalization method such as SoftMax function to get *A*. The regular attention layer is: $O = \hat{A} \times \gamma^T$, where $O \in \mathcal{R}^{L \times \hat{C}}$.

We exploit the non-local affinity matrix in final output layer of semantic segmentation tasks. To modify the original affinity, let $\phi, \theta, \gamma$ be identical mappings, and thus we can preserve the channel information, which represents the predicted probabilities. Firstly, we conduct a SoftMax normalization on score maps to make the channel information become probabilities. Then we conduct an element-wise square root operation on the embeddings Finally, we calculate affinity matrix *A*, which represents the similarities between every spatial locations.

To reduce the computational complexity of non-local affinity model, many previous work [11, 8] reduced the dimension of affinity matrix *A* and achieve significant performance. We followed the previous work [11] and adapt two sizes of down-sampling operations. Suppose the score map is of (C,W,H) size. We down sample the score map to (C,12,12) and (C,6,6), and then flatten them to (C,144), (C,36), and finally concatenate the two embeddings to get (C,180) embedding. The down-sampling location is aligned with label operations. Then we calculate the affinity matrix to get(180,180) sized $\widetilde{A}$,

$$\widetilde{A} = \sqrt{\phi^T} \times \sqrt{\theta} \qquad (5)$$

We associate it with the label affinity matrix by mean square error loss. The Affinity Regression loss is:

$$\mathcal{L}_{ar} = MSE(\widetilde{A}, M) \qquad (6)$$

Our total loss is:

$$\mathcal{L}_{total} = \mathcal{L}_{ce} + \lambda \mathcal{L}_{ar} \qquad (7)$$

We set $\lambda$ to be 0.1.

### 3.3. Mathematical Interpretation of Affinity Regression

In a general attention module, the affinity matrix *A* satisfies that:

$$A = \phi^T \times \theta \qquad (8)$$

specially $\phi = X, \theta = X$ in our settings, where *X* is the score map being down-sampled and channel-wise SoftMax normalized. Let $p_i$ denote the ith row vector of $X^T$ and $p_j$ denote the jth row vector of $X^T$.

We have:

$$\sum_{k=1}^{C} p_{ik} = 1, \sum_{k=1}^{C} p_{jk} = 1 \qquad (9)$$

$$\widetilde{A}_{ij} = \sum_{k=1}^{C} \sqrt{p_{ik} p_{jk}} \qquad (10)$$

According to Cauchy inequality:

$$\widetilde{A}_{ij}^2 = (\sum_{k=1}^{C} \sqrt{p_{ik} p_{jk}})^2 \leq \sum_{k=1}^{C} p_{ik} \sum_{k=1}^{C} p_{jk} = 1 \qquad (11)$$

and the inequality holds when $P_i$, $P_j$ have the same distributions. That is to say, when $A_{ij}$ regresses to 1, $P_i, P_j$ are forced to have the same distribution. When regresses to 0, $P_i$, $P_j$

are forced to have different distributions. It is obvious that
$\widetilde{A}_{ii} = \sum_{k=1}^{C} p_{ik} = 1$

Consider $p_i, p_j$ as two positive sequences $p_{i1}, p_{i2}...p_{iC}, p_{j1}, p_{j2}, p_{jC}$, according to rearrangement inequality : $\sum_{k=1}^{C} p_{ik}p_{jk}$ reaches maximum value when the two sequences $p_i$, $p_j$ have the same order, and when the two sequences have reversed order the summation reaches the minimum value.

In semantic segmentation tasks, the argmax of $p_i$ denotes the predicted labels.

Therefore, our regression operation can be a penalty. When the labels of two locations i, j are the same, the aligned vector $p_i$, $p_j$ of the prediction layer are penalized to have the same distribution. Especially, the argmax channels have to be the same. In other words, they predict the same class. When the labels are different, the penalty forces $p_i$, $p_j$ to have different distributions, especially on the argmax channels and some subargmax channels that mainly influence the summation. That is to say, they predict different classes.

We think this regression loss can be a structural supervision for semantic segmentation, which utilizes a binary supervision that cross-entropy loss cannot provide [14].

And , $\sqrt{p_{ij}} \in [0,1]$ promises the input of affinity layer being bounded, and thus the dot-product affinity mapping is a Lipschitz function [24]. A Lispchitz function has many good mathematical properties, including stable convergence. Fig- ure.3 (b) shows the rapid decline of affinity regression loss. Our model takes advantages of using pair-wise penalty and can be viewed as an auxiliary loss function which can be trained with segmentation networks simultaneously. And no extra parameters are added into the network.

## 4. EXPERIMENT

### 4.1. Dataset

NYU Depth Dataset V2 [26]. NYUv2 dataset is an indoor scene dataset which contains 1449 densely labeled pairs of aligned RGB and depth images, including 795 training images and 654 validation images. In our settings, we use 40 classes and do not make use of depth information in any way. Cityscapes Dataset [27]. Cityscapes is a dataset for semantic urban street scene understanding. 5000 high quality pixellevel densely annotated images that are divided into training, validation, and testing sets with 2975, 500, and 1525 images respectively. In our settings, we use 19 classes.

### 4.2. Experiment Settings

We firstly conduct our experiments on NYUv2. We use a light weighted refinenet [20] and PSPNet [3] as backbone networks, which perform well at this dataset. As is required in [20], experiments are limited on a single GPU without any

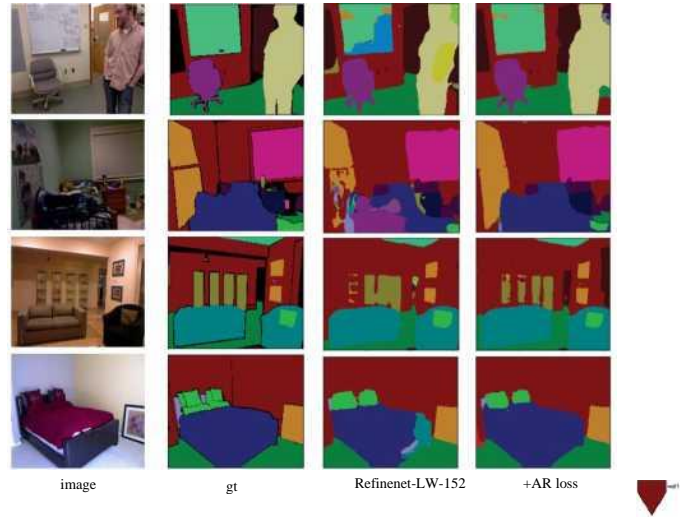

Figure.2. Segmentation results on NYUv2 validation dataset. We use Refinenet-LW-152 as baseline. From left to right: Input image, ground truth, baseline, baseline+AR loss. The right color bar represents color map.

TTA(testing time augmentation). In PSPNet experiment, 2 GPUs are used with batch size 6, and we follow [3] settings to use multi-scale testing method. Then we test our affinity regression loss on Cityscapes dataset. We adapt ICNet [21] as backbone, which is a sufficient real-time segmentation network. We follow the settings of [21], and our affinity regression loss is computed on the main output branch of IC- Net [21]. In every experiment, total training epochs are 200. Our experiments are based on PyTorch environment, with NVIDIA GeForce GTX 1080TI graphics cards

| Backbone | mIoU | +ARloss mIoU |
|---|---|---|
| Refinenet-LW-50 [20] | 41.7 | 42 |
| PSPNet-101 [3] | 42.6 | 43.3 |
| Refinenet-LW-152 [20] | 44.4 | 45.3 |

Table 1. Experiments on NYUv2 dataset

### 4.3. Analysis

We first train the original segmentation network to a comparable good accuracy and then add the regression penalty to continue training. In Figure.3, subfigure (b) shows that after we add the affinity regression loss to continue training, the loss quickly goes down.

From (a) in Figure.3, we can observe that our auxiliary loss stably increases the performance of the backbone segmentation network Even with less training epochs, our model can help backbones surpass the original peak performance. From Table 1 we can see that the regression penalty is workable in promoting segmentation accuracy.

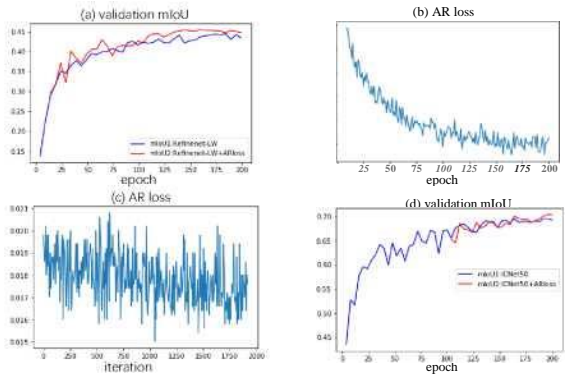

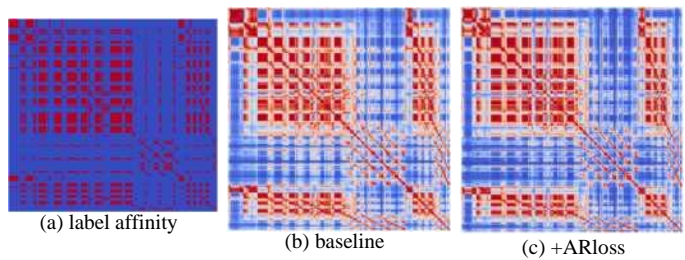

(a) label affinity  (b) baseline  (c) +ARloss

Figure.4. Heat map on NYUv2 validation set. Red color represents 1, while blue color represents 0. Color intensity indicates the closeness to 1 or 0. From left to right: label affinity, square root affinity of baseline, square root affinity of base-line+AR loss

Figure.3. Experiments on NYUv2 dataset and Cityscapes dataset. Subfigure (a)(b)(c) are based on Refinenet-LW-152 and subfigure (d) is based on ICNet-50.

As an implementation, we add our AR loss to fine-tune when the original network reaches its peak performance on validation set. The affinity regression loss decreases as iteration numbers increase. The curve (c) in Figure.3 has a tendency to go down in the fine-tuning process. We conduct a statistical hypothesis test to see if the downward trend is significant. In Table 2, we hypothetically use a linear regression model to fit the curve and then get the significance of parameters, where y represents the slope of the regression line. And the p-value is far smaller than 0.001, and thus we are over 99.9% confident that the downward trend is significant. And we calculate AR loss on validation set. We see a 11.8% decrease from 0.187 to 0.165.

To explore the generalization performance of our model, we use PSPNet-resnet101 [3] as segmentation backbone to continue experiments on NYUv2, and mIoU increases after finetuning, as seen in Table 1.

And to explore the generalization performance of our model on other datasets, we choose a real-time segmentation network ICNet [21], which performs well on Cityscapes dataset. From Figure.3 (d) we can see the validation performance of ICNet-50 after we add our model to train. mIoU increases stably in the final. On testing set, mIoU increases by 1.5%, as seen in Table 3.

| Method | mIoU |
|---|---|
| ICNet-50 [21] | 69.5 |
| ICNet-50+coarse data [21] | 70.5 |
| ICNet-50 [21]+ARloss | **71.0** |

Table 3. Results on Cityscapes testing set

| lm | Estimate | Std. Error | t value | Pr(>|t|) |
|---|---|---|---|---|
| Intercept | 1.864e-02 | 1.023e-04 | 182.210 | <2e-16 *** |
| y | -8.079e-07 | 9.227e-08 | -8.755 | <2e-16 *** |

Table 2. Significance Test

To visualize the decrease of AR loss on validation set, we use heat map to demonstrate the label affinity and the square root affinity matrix, as shown in Figure.4. Color intensity represents the similarity of position $(i, j)$. The bluer the color is, the more dissimilar $(p_i, p_j)$ is. The more red it is, the more similar $(p_i, p_j)$ is. The heat map is very intuitive. After finetuning the blue and red areas are both thicker. This visualization demonstrates that our model really guides the score map to learn label affinity.

Finally, we see the segmentation results on validation set of NYUv2 in Figure.2. We can observe that after adding our regression penalty, the prediction results are better for shape retention.

From the experiment results we can conclude that our theory works. The AR loss really guides score map to learn label affinity in the fine-tuning process. And the network learns to group the predicted probabilities of the same class and to separate the predicted probabilities of different classes, which really enhances the performance of segmentation networks.

## 5. CONCLUSION

We propose a novel paradigm of using label affinity as regression supervision in a binary form penalty for semantic segmentation, which is a classification-based task, and no extra parameters are added into networks. We demonstrate the effectiveness of our model on NYUv2 dataset and Cityscapes dataset.